%% file: main.tex
\documentclass[10pt,twocolumn,letterpaper]{article}

\usepackage{cvpr}              

\input{preamble}

\definecolor{cvprblue}{rgb}{0.21,0.49,0.74}
\usepackage[pagebackref,breaklinks,colorlinks,citecolor=cvprblue]{hyperref}
\def\paperID{7938} 
\def\confName{CVPR}
\def\confYear{2024}

\title{MCD: Diverse Large-Scale Multi-Campus Dataset for Robot Perception
}

\author{Thien-Minh Nguyen$^{1}$
\and Shenghai Yuan$^1$
\and Thien Hoang Nguyen$^1$
\and Pengyu Yin$^1$
\and Haozhi Cao$^1$
\and Lihua Xie$^1$
\and Maciej Wozniak$^2$
\and Patric Jensfelt$^2$
\and Marko Thiel$^3$
\and Justin Ziegenbein$^3$
\and Noel Blunder$^3$\\
\and $^1$ School of EEE, NTU, Singapore, $^2$ Division of RPL, KTH, Sweden, $^3$ ITL, TUHH, Germany
\and \shorturl{mcdviral.github.io}
\vspace{-0.3cm}
}

\begin{document}

\twocolumn[{
\renewcommand\twocolumn[1][]{#1}
\maketitle
\begin{center}
    \centering
    \vspace{-0.6cm}
    \captionsetup{type=figure}
    \includegraphics[width=0.9\linewidth]{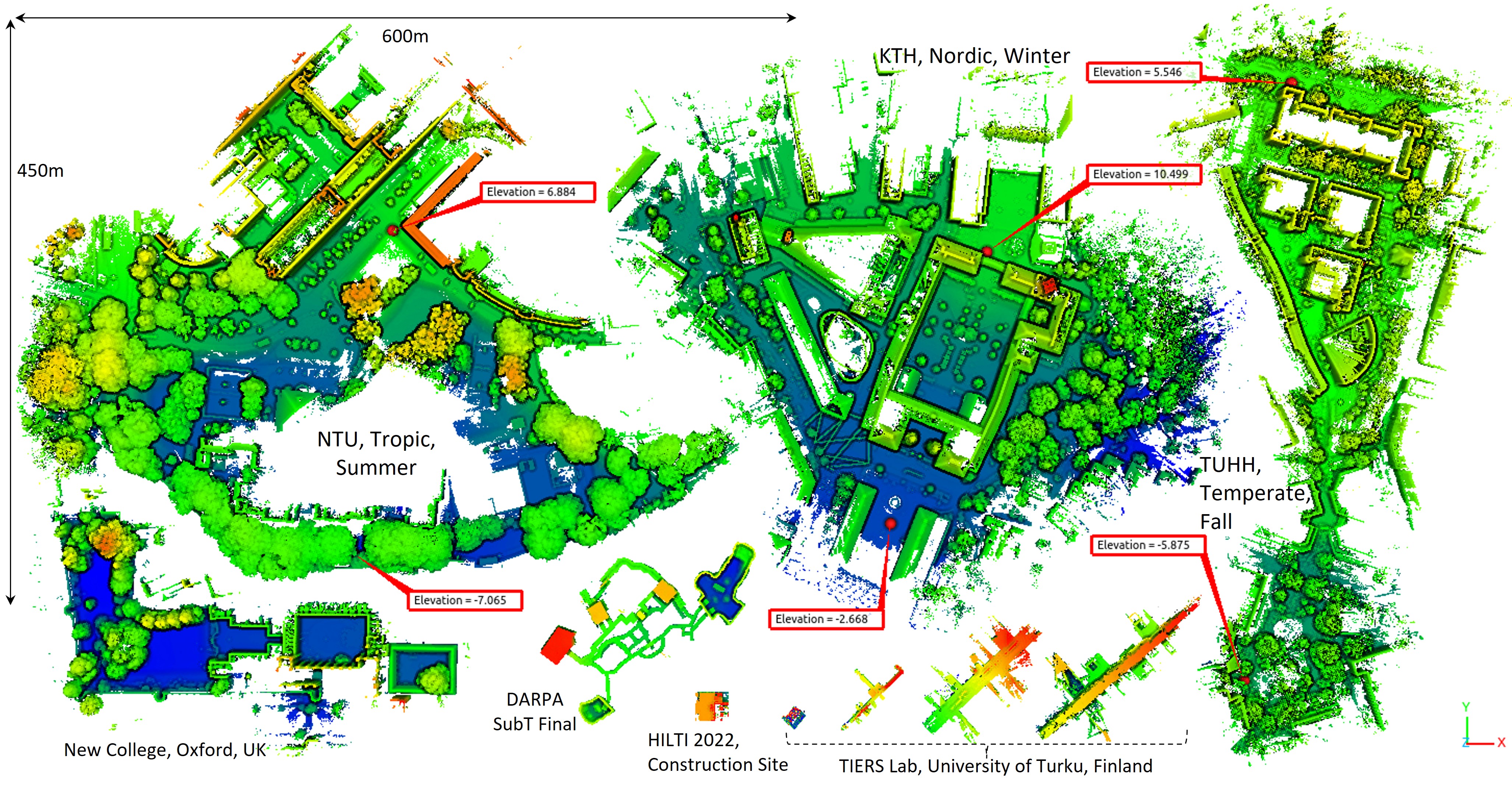}
    \captionof{figure}{Up-to-scale comparision of survey-grade prior maps in MCD with other works.}
    \label{fig: prior maps}
\end{center}%
}]

\maketitle

\begin{abstract}
\vspace{-0.25cm}
Perception plays a crucial role in various robot applications.
However, existing well-annotated datasets are biased towards autonomous driving scenarios, while unlabelled SLAM datasets are quickly over-fitted, and often lack environment and domain variations. To expand the frontier of these fields, we introduce a comprehensive dataset named MCD (Multi-Campus Dataset), featuring a wide range of sensing modalities, high-accuracy ground truth, and diverse challenging environments across three Eurasian university campuses. MCD comprises both CCS (Classical Cylindrical Spinning) and NRE (Non-Repetitive Epicyclic) lidars, high-quality IMUs (Inertial Measurement Units), cameras, and UWB (Ultra-WideBand) sensors. Furthermore, in a pioneering effort, we introduce semantic annotations of 29 classes over 59k sparse NRE lidar scans across three domains, thus providing a novel challenge to existing semantic segmentation research upon this largely unexplored lidar modality.
Finally, we propose, for the first time to the best of our knowledge, continuous-time ground truth based on optimization-based registration of lidar-inertial data on large survey-grade prior maps, which are also publicly released, each several times the size of existing ones.
We conduct a rigorous evaluation of numerous state-of-the-art algorithms on MCD, report their performance, and highlight the challenges awaiting solutions from the research community.
\end{abstract}

\vspace{-0.5cm}

\section{Introduction}

Understanding the world and estimating egomotion are fundamental in computer vision and robotics, enhancing applications such as autonomous driving, logistics, delivery, and AR/VR (Augmented/Virtual Reality). Over the years, public datasets have played a crucial role in promoting broader participation in the research of environment perception and egomotion estimation, especially for those with limited resources. However, we find that existing annotated multi-modality large-scale datasets are biased toward autonomous driving scenarios, and rely heavily on costly classical lidars and cameras. These datasets are often significantly detached from real-world applications. For instance, cameras can give rise to privacy concerns in many countries and are susceptible to variations in lighting conditions and background noises. The high cost of dense CCS lidars often hinders their widespread adoption.

Recently, newer modalities for egomotion estimation and environment perception have surfaced, featuring low-cost NRE (Non-Repetitive Epicyclic) lidar and UWB (Ultra-WideBand) technology. These innovations address issues like cost, illumination, privacy, and robustness, but they also present new challenges, such as scan sparsity and limited field of view of NRE lidar, and NLOS (Non-Line-Of-Sight) observations by UWB. These factors complicate environment perception tasks, but also open up new research opportunities.

To facilitate this research, we introduce MCD, a multi-campus dataset designed to address various robotics perception challenges. Our primary goal is to foster the development of next-generation, cost-effective robots, AI (Artificial Intelligence), and AR/VR systems capable of working in diverse environments, not limited to public roads.
\textit{The MCD boasts an extensive collection of 18 sequences, over 200k lidar scans, 1500k camera frames, and high-frequency IMU (Inertial Measurement Unit) and UWB data}. Notably, \textit{the dataset includes point-wise annotations for typical outdoor and indoor objects}, making it one of the pioneering works to provide such annotations for affordable NRE lidar systems. Moreover, we also propose a novel \textit{continuous-time ground truth data}, which demonstrates superior accuracy when compared to other datasets of similar scales.
The contributions and features of MCD are listed below.

\noindent \textbf{Multiple Sensing Modalities}. MCD includes both classical rotating lidar and newer MEMS NRE lidar, as well as traditional cameras with different baselines.
Additionally, it incorporates useful modalities like IMUs and UWBs, which were not present in widely used benchmarks like KITTI \cite{Geiger2013IJRR} \cite{behley2019semantickitti}, EuRoC  \cite{burri2016euroc}, Newer College \cite{ramezani2020newer}. The manual for using these modalities can be found at {\footnotesize\shorturl{mcdviral.github.io/UserManual.html}}.

\begin{figure}
    \centering
    \includegraphics[width=\linewidth]{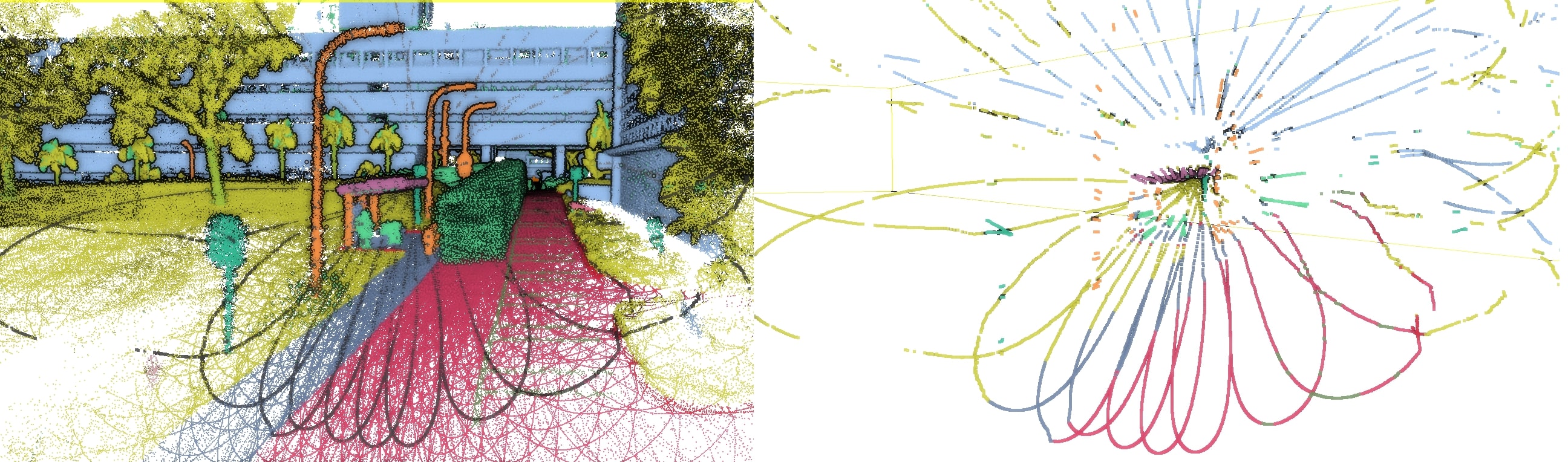}
    \caption{Annotated NRE lidar scan motion concatenated over 20s period (left) and a single scan (right). This modality remains untested in semantic segmentation research to the best of our knowledge.}
    \label{fig: livox semantics}
    \vspace{-0.25cm}
\end{figure}

\noindent \textbf{Pointwise Annotations for NRE Points Clouds}. Low-cost, accurate NRE lidars have gained attention but are underexplored in semantic segmentation. Their sparsity and irregular scanning pose challenges for both learning methods and human annotators, as depicted in \ref{fig: livox semantics}. Despite this, we have carefully annotated 59k lidar scans, surpassing  SemanticKITTI \cite{behley2019semantickitti}  and Nuscenes \cite{caesar2020nuscenes}. MCD is the first extensive dataset with semantic annotations for NRE lidars.
The NRE scans and preview of annotated sequences can be found at {\footnotesize\shorturl{mcdviral.github.io/AnnotatedLidar}}.

\begin{figure}
    \centering
    
    \includegraphics[width=0.32\linewidth]{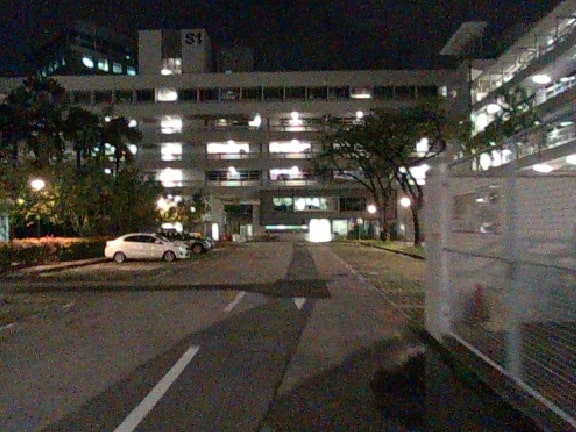}
    \includegraphics[width=0.32\linewidth]{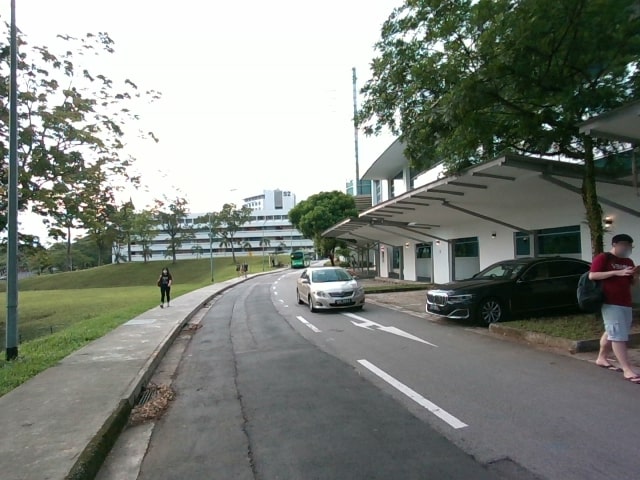}
    \includegraphics[width=0.32\linewidth]{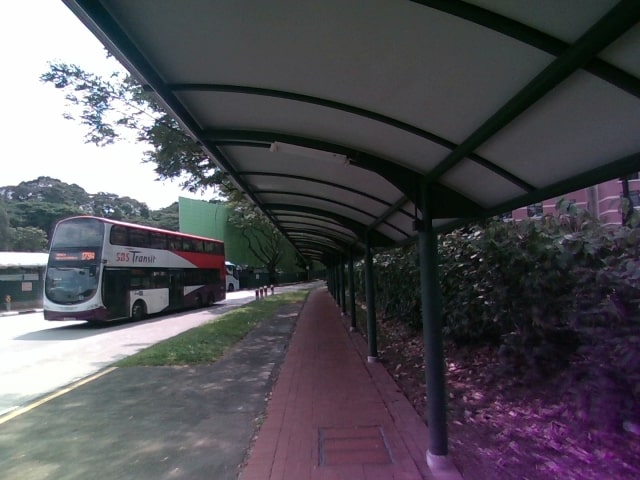}

    \includegraphics[width=0.32\linewidth]{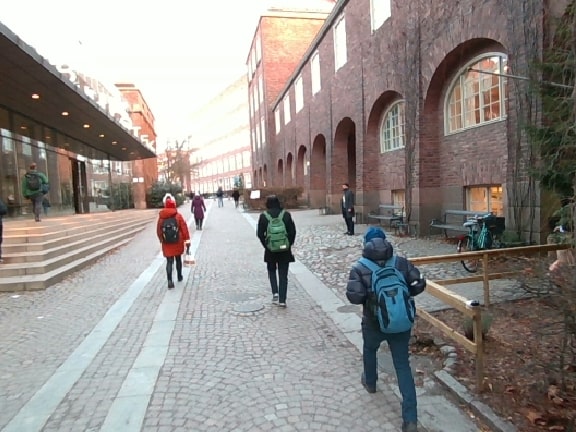}
    \includegraphics[width=0.32\linewidth]{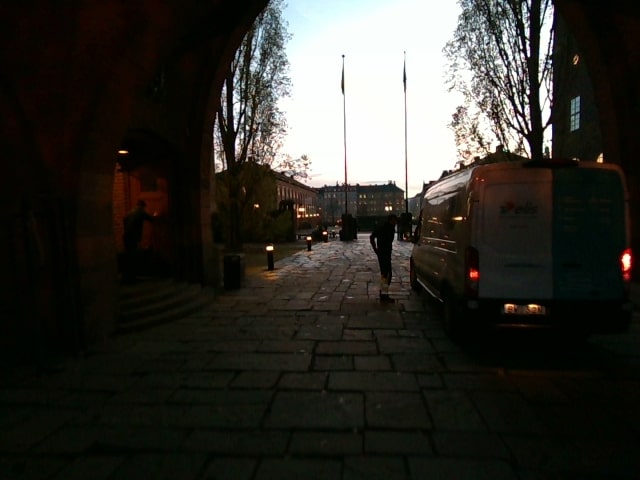}
    \includegraphics[width=0.32\linewidth]{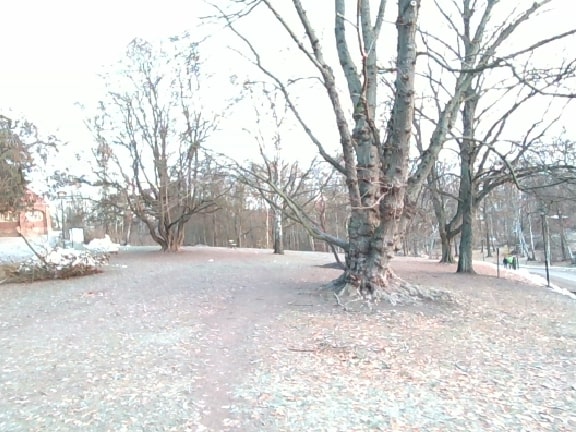}

    \includegraphics[width=0.32\linewidth]{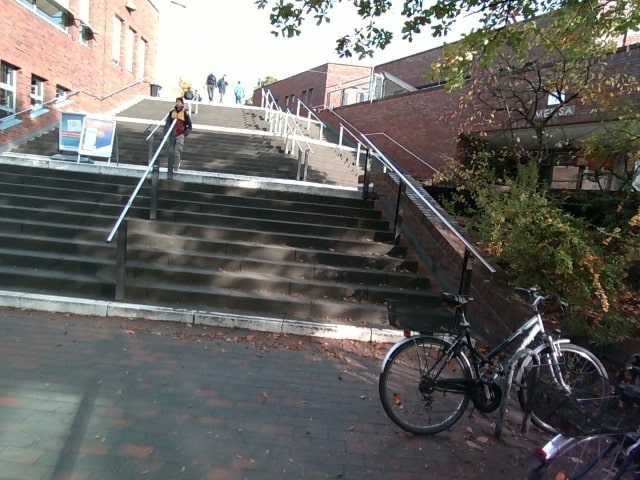}
    \includegraphics[width=0.32\linewidth]{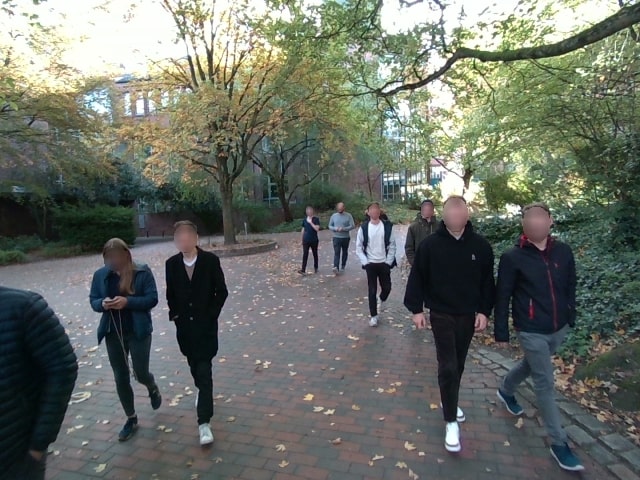}
    \includegraphics[width=0.32\linewidth]{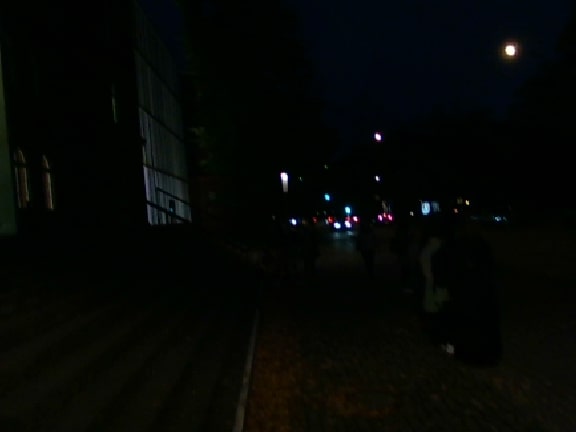}
    
    \caption{Example camera views at different campuses in MCD, with various terrains, backgrounds and lighting conditions. All faces are anonymized in compliance with local regulations.}
    \label{fig: elevations}
\end{figure}

\noindent \textbf{Wider Domain Coverage}. We collect data from three university campuses across Eurasia, covering a broader latitude range than most autonomous driving datasets \cite{Geiger2013IJRR} \cite{behley2019semantickitti} \cite{carlevaris2016university} \cite{caesar2020nuscenes} \cite{sun2020waymoscalability}, as shown in \ref{fig: elevations}. This diversity leads to notable variations in feature prior distribution, providing challenges for general robot learning and comprehension.

\noindent \textbf{Continuous-time Ground Truth}.
Accurate ground truth is vital for SLAM and localization studies, especially in larger environments. Existing datasets offer only discrete-time ground truth poses or processed lidar scans to reduce motion distortion, limiting research possibilities. To address these issues, we propose continuous-time ground truth based on survey-grade prior map registration. This continuous-time ground truth allows arbitrary time and density sampling. It is an essential feature for high FPS gaming AR/VR devices.
Detailed explanations and usage of ground truth can be found at {\footnotesize\shorturl{mcdviral.github.io/Groundtruth}}.

\begin{figure}
    \centering
    \includegraphics[width=\linewidth]{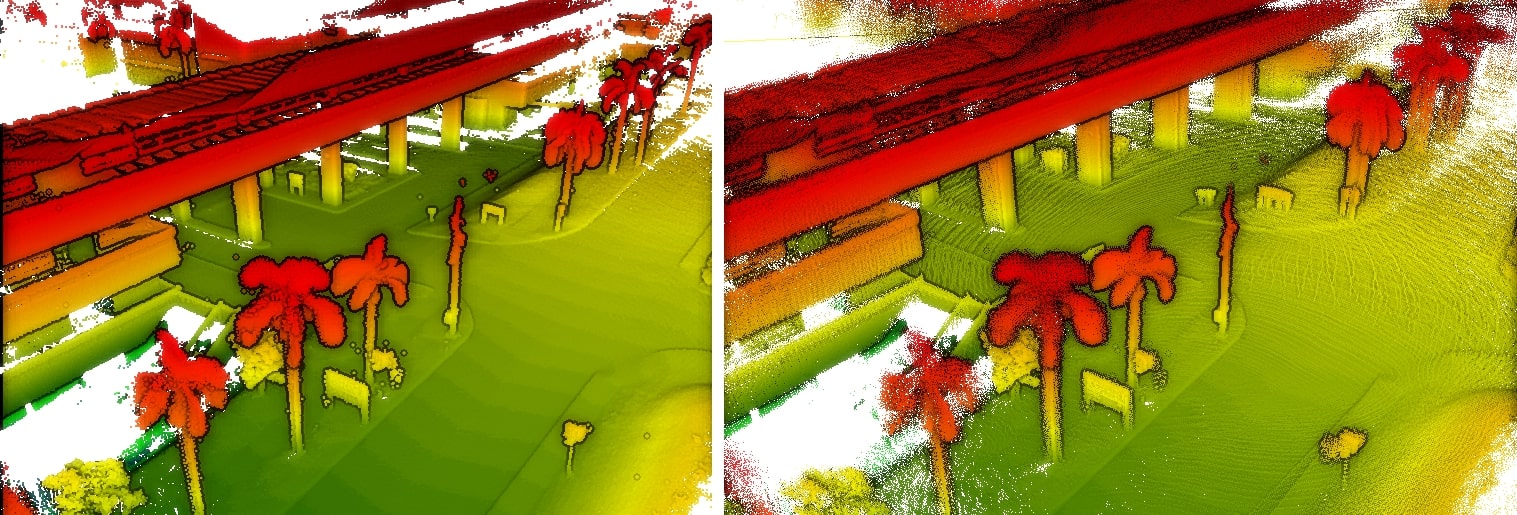}
    \caption{The lidar point clouds with (left) and without (right) motion undistortion.}
    \label{fig: motion}
\end{figure}

\begin{figure}
    \centering
    \includegraphics[width=0.48\linewidth]{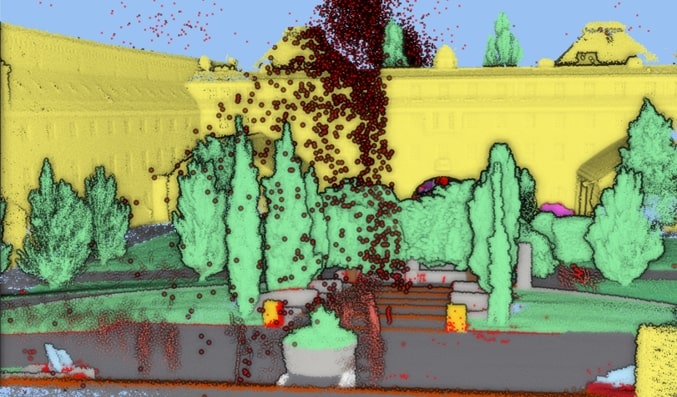}
    \includegraphics[width=0.48\linewidth]{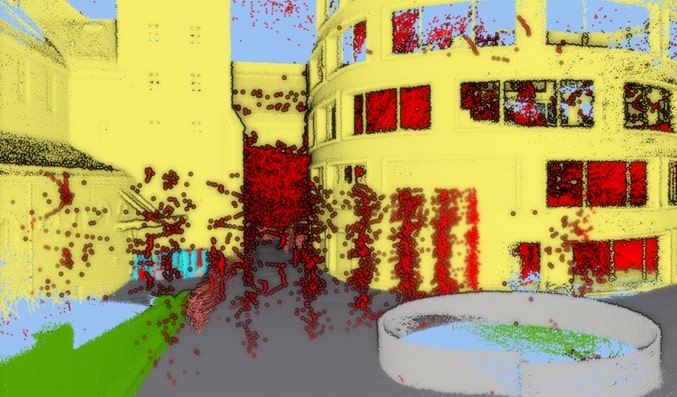}
    \caption{The sunglare and reflection outliers in the lidar scans are uncommon in lidar datasets. Significant effort were needed to segment out these outlier points. }
    \label{fig: noise}
    \vspace{-0.25cm}
\end{figure}

\noindent \textbf{Embrace the Challenges in Perception}.
We believe understanding motion distortion is critical in real-world end-to-end training scenarios and should not be obscured, as it has significant implications for research. Researchers can choose to retain or remove motion distortion from raw lidar point clouds, as shown in \Fig{fig: motion}.
MCD tackles other challenges like extreme lighting, glass reflection, and solar interference, as illustrated in \Fig{fig: noise}, which have led to real-world accidents due to developers overlooking them. However, these issues are often inadequately addressed in existing datasets. We incorporate those noise classes in annotations and images as part of our efforts to train robotics systems to handle such corner cases.

\noindent \textbf{Extensive Benchmarks}.
To facilitate and advance future research endeavors, we provide the dataset, data-loading scripts, and benchmarking instructions. Furthermore, We conducted benchmarking for both semantic segmentation and SLAM studies, providing in-depth understanding and analysis of each method. These benchmarks highlight the considerable room for improvement and the opportunities for innovation in each field. The instructions can be found at {\footnotesize\shorturl{mcdviral.github.io/QuickUse}}.


\section{Related Works}
The MCD draws inspiration and insights from several preceding works, each representing a unique milestone in developing autonomous systems. A comprehensive table that compares existing multi-modal perception datasets with MCD can be found in the supplementary material. Below, we shall provide a short synopsis of the evolution of multi-modal datasets in the last decade.

Released more than a decade ago, the KITTI dataset \cite{Geiger2012CVPR, Geiger2013IJRR} stands out as a classic and is widely recognized in autonomous driving research.
The KITTI dataset and its subsequent publications have enabled the evolution of numerous studies in areas such as stereo matching, optical flow \cite{Menze2015CVPR}, depth perception \cite{Uhrig2017THREEDV}, object detection \cite{Fritsch2013ITSC}, tracking \cite{Luiten2020IJCV}, and semantic segmentation \cite{Alhaija2018IJCV}. Later, SemanticKITTI \cite{behley2019semantickitti} was introduced to offer semantic annotations for all sequences within the KITTI Odometry Benchmark.
However, the key limitation of the KITTI dataset lies in the absence of IMU data and low diversity in terms of environment and terrain, thereby impeding comprehensive research in domains like domain adaptation and inertial-based SLAM. Numerous autonomous driving datasets, such as Cityscapes \cite{cordts2016cityscapes}, Urban Loco \cite{wen2020urbanloco}, NuScene \cite{caesar2020nuscenes}, Waymo \cite{sun2020waymoscalability}, ApolloScape \cite{huang2018apolloscape}, A*3D \cite{pham20203d} and H3D \cite{patil2019h3d}, have emerged recently, each offering data across various locations. These datasets predominantly focus on road-centric driver views, thus lacking diversity in the environment. They often fail to deliver millimeter-level ground truth precision due to GPS/INS fusion limitations.

NCLT \cite{carlevaris2016university} aims to enhance diversity by providing a long-term, domain-shifted dataset that extends over a year. Nonetheless, this dataset still has challenges related to ground truth accuracy and is restricted to a single location.
The challenge of geographical limitation is a common problem in dataset development. For example,  KAIST Urban \cite{jeong2019complex}, TorontoCity \cite{wang2016torontocity}, and Oxford RobotCar \cite{maddern20171}, also have similar limitations, operating exclusively within one city and lack millimeter-level ground truth.

To address perception and scene understanding in a general environment with higher ground truth accuracy, the Newer College Dataset \cite{ramezani2020newer} was introduced, utilizing a ground truth constructed from scanned survey maps for both indoor and outdoor scenes. However, the ground truth data contain noticeable errors due to failures in the ICP (Iterative Closest Point) matching of the lidar scan with the prior map. This issue primarily arose due to static maps and moving scans collected at different times of the year, resulting in noticeable differences in vegetation growth. Additionally, this dataset lacks ground truth in semantic labels and covers very small areas, significantly limiting its impact on AI and SLAM research. The Hilti SLAM Challenge \cite{hilti2021,zhang2022hilti} also employ a comparable method for ground truth generation. However, their keyframe poses are prone to overfitting when using reinforcement learning-like hyperparameter tuning \cite{lim2023adalio} for scoring. Additionally, these datasets do not include any annotations.

Most well-established datasets struggle with providing high-accuracy position ground truth, and only a few studies attempt to address these issues. The UAV flying dataset named NTU VIRAL \cite{nguyen2021ntuviral} aims to overcome this accuracy limitation by incorporating two lidars, stereo vision, inertial sensors, and UWB on a single drone, supplemented with survey-grade tracking tools. This approach draws significant inspiration from the EUROC dataset \cite{burri2016euroc}. However, these datasets still suffer from limited domain coverage and lack of annotations. Additionally, they require a line of sight to ground survey measuring stations, which makes them unsuitable for large-scale dataset applications.

Simulated datasets such as CARLA \cite{dosovitskiy2017carla} and TartanAir \cite{wang2020tartanair} offer a wide range of environments along with absolute ground truth data. However, they exhibit a noticeable perceptual domain shift due to their simulator models and are limited by their environmental settings. Furthermore, their inability to completely replicate real-world physics and dynamics, such as sun glare and mirror reflections, often limits their relevance in practical real-world scenarios.

\begin{figure}
    \centering
    \includegraphics[width=\linewidth]{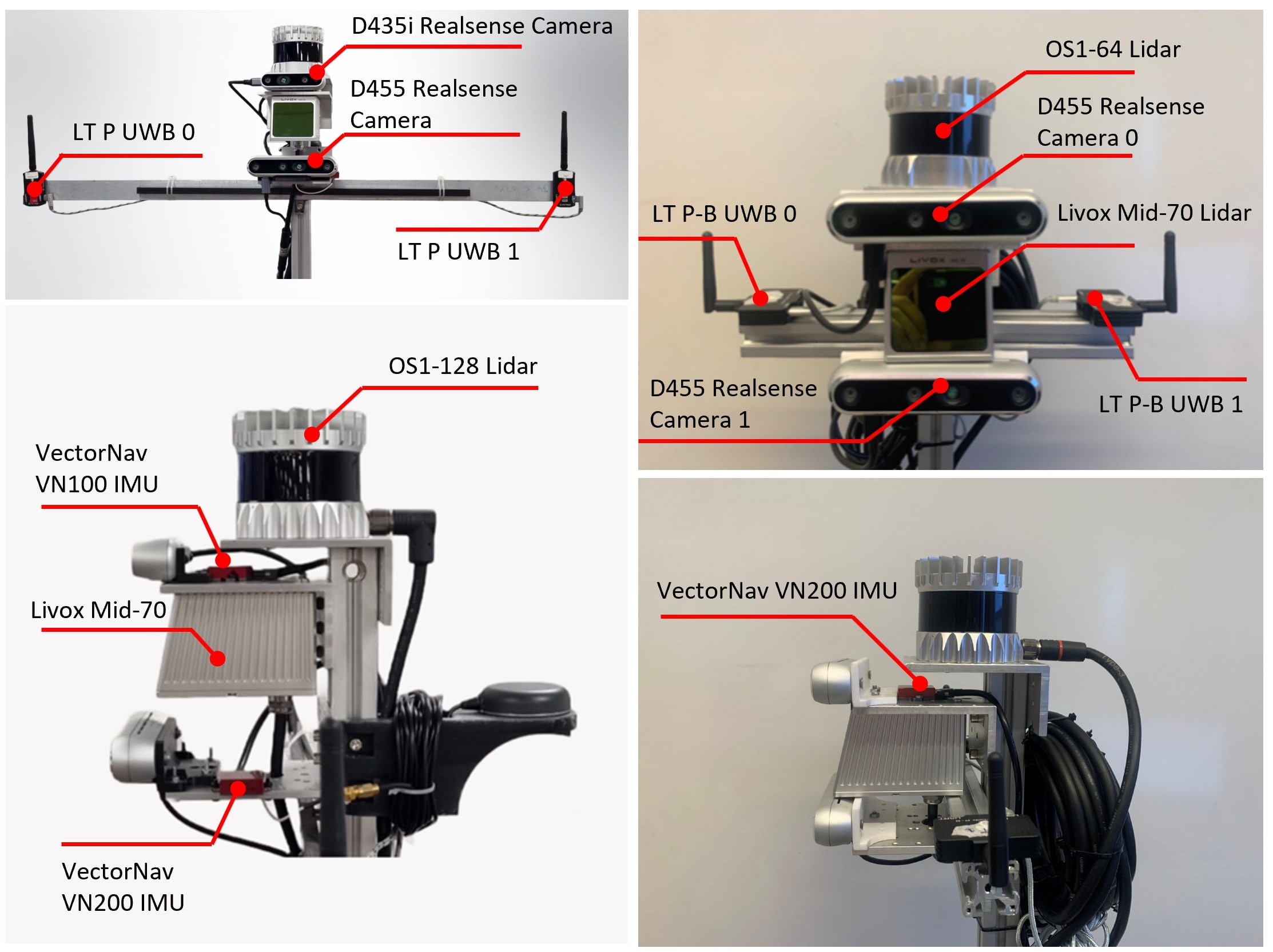}
    \caption{Sensing modalities on the sensor suites: ATV (left) and HHS (right).}
    \label{fig: sensor suite}
    \vspace{-0.25cm}
\end{figure}

\section{Dataset Statistical Analysis}

\subsection{Sequence Characteristics}

The MCD dataset consists of 18 sequences, with six sequences from each campus. Within each campus, three sequences are captured during the day and three at night. Table \ref{tab: sequences} lists the names of these sequences and some of their characteristics. Visualization of the paths can be viewed at {\footnotesize\shorturl{mcdviral.github.io/Download}}.

The route of each sequence is decided with deliberation. For each campus, we start by collecting one sequence on a long path covering most of the environment, so-called the main sequences. Then, we collect two shorter sequences that largely overlap with parts of the main sequence, though each still features some exclusive tracks. This arrangement facilitates researchers in analyzing the limitations of SLAM methods within the visual and geometrical contexts of the environment. Moreover, we collect another set of sequences in the nighttime, mainly following the same path as daytime sequences. 
Hence, a wide range of algorithms can be evaluated for their robustness in the context of visual methods under challenging lighting conditions.

\begin{figure*}
    \centering
    
    \includegraphics[width=0.32\linewidth]{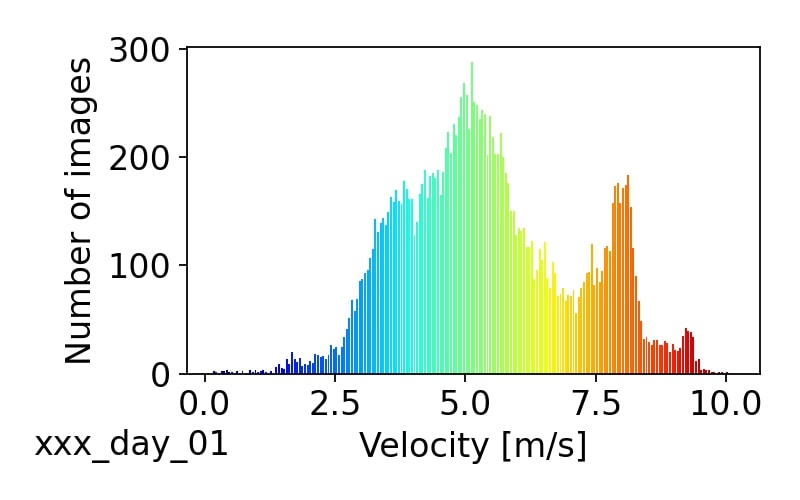}
    \includegraphics[width=0.32\linewidth]{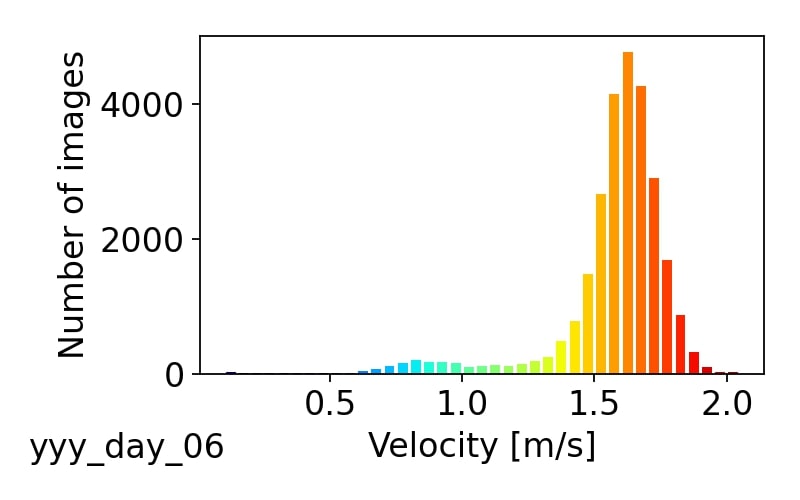}
    \includegraphics[width=0.32\linewidth]{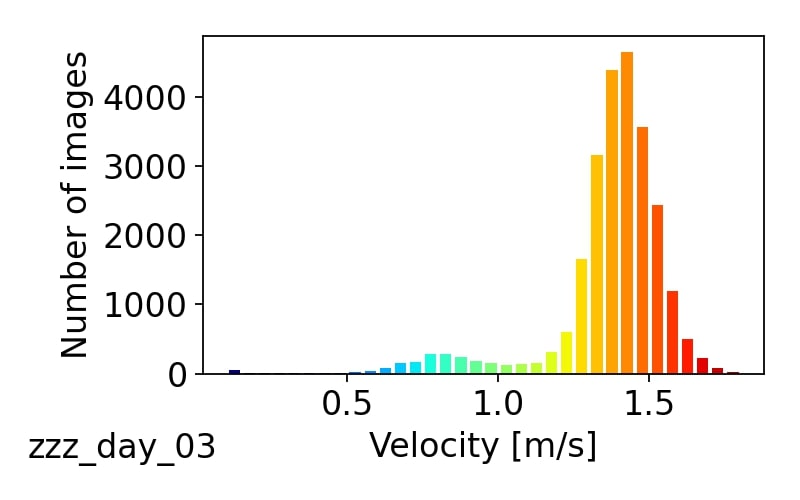}
    \caption{Velocity histograms of the images under topic /d455b/color/image\_raw in the three main sequences.}
    \label{fig: velocity hist}
    \vspace{-0.25cm}
\end{figure*}

Fig. \ref{fig: velocity hist} shows the velocity histograms of images captured under a camera topic across three main sequences. The maximum velocity can approach 10m/s in the NTU sequences, which is almost five times that of the KTH and TUHH sequences. This difference is due to ntu sequences being collected from the ATV (All Terrian Vehicle) and the others from Handheld Setup (HHS), as shown in Fig. \ref{fig: sensor suite}.
These velocity profiles will allow users to test their algorithms against low or high-speed conditions. This issue is also discussed in Sec. \ref{sec: lidar-inertial slam}

\def\toprule{\hline\hline}
\def\midrule{\hline}
\def\bottomrule{\hline\hline}

\begin{table}
\centering
\renewcommand{\arraystretch}{1.125}

\begin{threeparttable}

\caption{Sequence names, their duration (Dur.), traversed distance (Dist.), maximum and median velocity (max(V) and med(V)). The sequences in \textbf{bold} have their livox point clouds annotated.} \label{tab: sequences}

\begin{tabular}{lccccc}
\toprule
\mr{2}{*}{Seq. Name}  &
                           Dur.  &     Dist.&   max(V) &    med(V) \\
                        &(mm:ss) &     (m)  &   (km/h) &    (km/h)  \\
\midrule
   \bf{ntu\_day\_01}   &   10:02  &     3198 &   36.70  &    18.69 \\
   \bf{ntu\_day\_02}   &   03:49  &      642 &   21.02  &    11.37 \\
   \bf{ntu\_day\_10}   &   05:25  &     1783 &   33.05  &    20.02 \\
   ntu\_night\_04      &   04:56  &     1459 &   27.18  &    18.27 \\
   ntu\_night\_08      &   07:47  &     2421 &   34.38  &    19.01 \\
   \bf{ntu\_night\_13} &   03:54  &     1231 &   29.40  &    20.27 \\
\midrule
   \bf{kth\_day\_06}   &   14:51  &     1403 &    7.54  &     5.82 \\
   \bf{kth\_day\_09}   &   12:47  &     1076 &    7.82  &     5.25 \\
   kth\_day\_10        &   10:15  &      920 &    8.12  &     5.82 \\
   kth\_night\_01      &   16:09  &     1416 &    7.53  &     5.49 \\
   kth\_night\_04      &   12:26  &     1052 &    7.15  &     5.51 \\
   \bf{kth\_night\_05} &   11:05  &      919 &    7.37  &     5.52 \\
\midrule
  \bf{tuhh\_day\_02}    &   08:20  &      749 &    8.02  &     5.95 \\
  \bf{tuhh\_day\_03}    &   13:59  &     1137 &    6.51  &     5.05 \\
  tuhh\_day\_04         &   03:08  &      297 &    7.77  &     6.32 \\
  tuhh\_night\_07       &   07:24  &      742 &    8.55  &     6.57 \\
  \bf{tuhh\_night\_08}  &   11:49  &     1128 &    7.73  &     6.10 \\
  \bf{tuhh\_night\_09}  &   03:05  &      290 &    7.66  &     6.21 \\
\bottomrule
\end{tabular}

\end{threeparttable}
\end{table}

\subsection{Semantic Annotations}

\begin{figure*}
    \centering
    \includegraphics[width=0.95\linewidth]{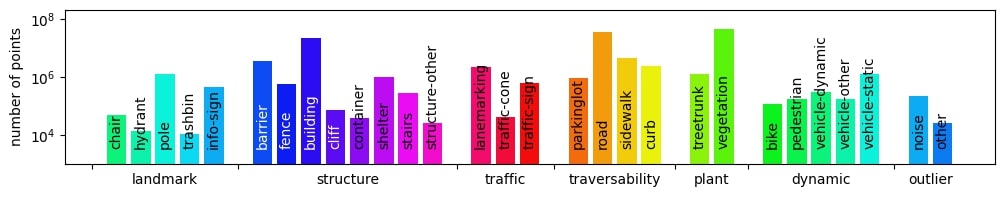}
    \includegraphics[width=0.95\linewidth]{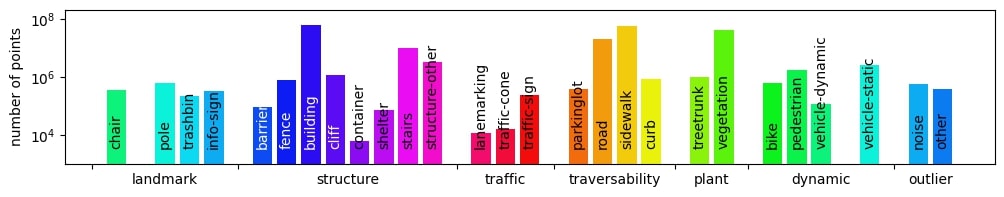}
    \includegraphics[width=0.95\linewidth]{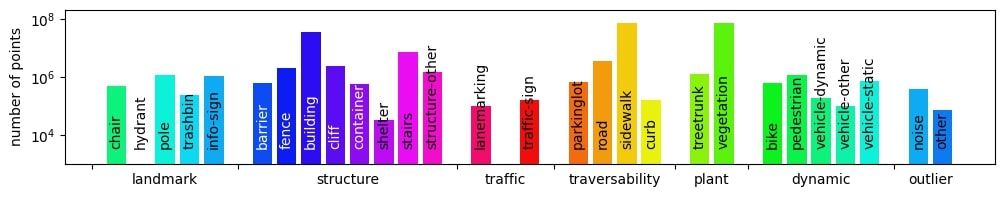}
    \caption{Distribution of the semantic classes in the three campuses.}
    \label{fig: classes}
\end{figure*}

A total of 29 semantic classes of 6 groups were applied on the Livox point clouds of 11 sequences.
From \Fig{fig: classes}, we can notice the domain shift of the three environments. For example, the number of points in poles, info-sign, vegetation, and buildings is roughly the same among the campuses. However, the data points for chairs and trash bins in kth and tuhh are an order of magnitude larger than those in the ntu campus. This discrepancy is primarily attributed to the geographical and climate differences.
NTU, being situated in a low-latitude region with a tropical climate, has fewer people sitting outdoor, in contrast to the other two campuses located in mid to high-latitude areas where the weather is more temperate. Additionally, NTU has more hydrants, likely influenced by the same climatic factors. The scarcity of bikes and riders in NTU is due to the combination of its harsh tropical weather and legal restrictions on e-scooters. Similar contrasts are observed in the structure group (stairs vs barriers) due to different styles of structures. The traffic cone reflects the different levels of construction, whereas the road/sidewalk ratio also reflects urban planning priorities. Differences in lane markings imply varying policies on pedestrian and vehicle regulations among the three campuses, with the Asia campus being more stringent and the Europe two campus being more lenient on traffic rules. While parking lot point distribution is similar, the Asia campus car park is more centralized, while the Europe campus car park is more distributed along the roadside. These domain shifts lead to unique lidar point cloud prior distributions across the campuses. Ignoring prior differences can impact prediction accuracy significantly, which is discussed in Sec. \ref{sec: semantic segmentation}.
\vspace{-0.75cm}

\section{Survey Map Continuous-Time Registration} \label{sec: ground truth}

In previous works, such as the Newer College Dataset (NCD) \cite{ramezani2020newer}, Multi-lidar Benchmark \cite{sier2023benchmark}, and Hilti-SLAM-Challenge\cite{zhang2022hilti}, simple ICP matching of SLAM-based undistorted point clouds with a prior map was used to provide discrete-time ground truth. However, we notice that these point clouds were generated through imperfect SLAM processes, introducing an inherent observational bias. This bias can then propagate, leading to a potential ground truth bias. In the case of NCD \cite{ramezani2020newer}, we can identify instances of ground truth bias, primarily stemming from ICP matching failures in certain sequences. These failures frequently arise due to the six-month time gap between the prior map data collection and the sequence data collection, resulting in observable discrepancies in the scanned data consistency, largely attributable to vegetation growth. 
In contrast, MCD data collection at each site spans less than one month, contributing to better ground truth consistency. 
Unlike the simple discrete-time ICP process, our method directly registers the raw point cloud with the prior map through a continuous-time optimization scheme that explicitly considers the deskew process during optimization. The cost function consists of pose priors, lidar-matching, and IMU factors as follows
\begin{align} \label{eq: spline optimization}
    &f\ll( \{ \Z_{t_n} \}, \hat{\tf}_t, \hat{\bias}\rr) = 
    \sum f_\ssp + \sum f_\ssl \nonumber \\
    &\qquad\qquad\qquad\qquad\qquad + \sum f_\ssg + \sum f_\ssa,
\end{align}
where $\{ \Z_{t_n} \}$ is the set of all lidar and inertial measurements, $\hat{\tf}_t$ is the continuous time trajectory estimate based on B-spline formulation \cite{sommer2020efficient}, $\hat{\bias}$ is the IMU bias estimate. The individual factors in \eqref{eq: spline optimization} are defined as
\begin{align*}
    f_\ssp  &=   \norm{\Log(\bar{\rot}_{t_k}^{-1} \hat{\rot}(t_k))
                      }^2_{\cov_\rot}
               + \Big\|{\hat{\pos}(t_k) - \bar{\pos}_{t_k}
                 \Big\|}^2_{\cov_\pos},
    \\
    f_\ssl  &=   \norm{  \n_\text{pm}^\top[\hat{\rot}(t_j){}^{t_i}\f
                       + \hat{\pos}(t_i)]
                       - \mu_\text{pm}}^2_{\cov_\ssl},
    \\
    f_\ssg  &= \norm{  \hat{\rot}^{-1}(t_j){}^\fr{W}\hat{\omega}(t_j)
                     + \hat{\bias}_\ssg
                     - {}^{\fr{B}_{t_j}}\breve{\omega}}^2_{\cov_\ssg},
    \\
    f_\ssa  &= \norm{ \hat{\rot}^{-1}(t_j)\ll[{}^\fr{W}\hat{a}(t_j) + \vbf{g}\rr]
                    + \hat{\bias}_\ssa - {}^{\fr{B}_{t_j}}\breve{a}}^2_{\cov_\ssa},
\end{align*}
where $\cov_\rot$, $\cov_\pos$, $\cov_\ssl$, $\cov_\ssg$, $\cov_\ssa$ are the covariance of the measurement error, $\n_\text{pm}$ and $\mu_\text{pm}$ are the coefficients of the plane contained within the voxel associated with the lidar point ${}^{t_i}\f$, $\vbf{g}$ is the gravity constant vector, and ${}^\fr{W}\hat{\omega}(t_j)$, ${}^\fr{W}\hat{a}(t_j)$ are the angular velocity and acceleration estimates, whereas ${}^{\fB_{t_j}}\hat{\omega}$, ${}^{\fB_{t_j}}\hat{a}$ are the angular velocity and acceleration measurements themselves.

\section{Benchmarking on SOTA algorithms}

In this section, we conduct experiments on SOTA (state-of-the-art) SLAM methods with MCD to demonstrate its complexity as well as find out the capability of these methods.

\def\toprule{\hline\hline}
\def\midrule{\hline}
\def\bottomrule{\hline\hline}

\begin{table*}
\centering
\renewcommand{\arraystretch}{1.125}

\begin{threeparttable}

\caption{ATE$^*$ of SOTA SLAM methods.} \label{tab: ATE}

\begin{tabular}{lllll|lll|lll}
\toprule
     sequence   &LIOSAM      &FLIO        &DLIO        &SLICT       &VINS         &KMR          &O.V.         & RO          & VRO \\
\midrule
   ntu\_day\_01 &      6.658 & \tb{1.510} &      1.925 & \ul{1.890} & \ul{15.510} &           x & \tb{10.605} & \ul{14.965} &  \tb{9.430} \\
   ntu\_day\_02 & \ul{0.227} &      0.272 &      0.636 & \tb{0.168} &  \ul{4.373} &           x &  \tb{1.683} &  \ul{9.175} &  \tb{3.408} \\
   ntu\_day\_10 &      2.845 & \ul{2.084} &      3.052 & \tb{1.429} & \tb{16.021} &           x & \ul{17.604} & \ul{14.156} &  \tb{7.360} \\
 ntu\_night\_04 & \ul{1.134} &      1.599 &      2.373 & \tb{1.002} &  \tb{5.262} &           x &  \ul{5.890} & \ul{25.584} & \tb{13.628} \\
 ntu\_night\_08 &      3.672 & \ul{1.425} &      2.056 & \tb{0.822} &           x &           x & \tb{13.662} & \tb{12.993} &           x \\
 ntu\_night\_13 &      0.923 & \ul{0.903} &      1.928 & \tb{0.574} &           x &           x & \tb{21.794} & \tb{10.777} & \ul{49.067} \\
   kth\_day\_06 & \tb{0.524} &      1.005 & \ul{0.562} &      0.633 & \tb{20.045} &      83.393 & \ul{72.445} & \ul{30.771} & \tb{15.490} \\
   kth\_day\_09 & \tb{0.235} &      0.733 &      0.326 & \ul{0.262} &           x &           x & \tb{22.494} & \ul{26.740} & \tb{17.366} \\
   kth\_day\_10 & \tb{0.255} &      2.176 & \ul{0.665} &      0.737 &      21.022 & \ul{17.092} & \tb{10.510} & \ul{26.691} & \tb{11.550} \\
 kth\_night\_01 &     12.240 &      1.040 & \tb{0.414} & \ul{0.540} &           x &           x &           x & \tb{27.284} & \ul{42.581} \\
 kth\_night\_04 & \tb{0.234} &      0.567 & \ul{0.376} &      0.441 & \ul{66.256} & \tb{14.947} &           x & \ul{27.678} & \tb{24.719} \\
 kth\_night\_05 & \tb{0.298} &      2.158 &      0.903 & \ul{0.855} &           x &           x &           x & \tb{30.421} & \ul{50.539} \\
  tuhh\_day\_02 & \ul{0.256} &      0.273 &      0.283 & \tb{0.236} &      16.349 &  \tb{3.735} &  \ul{6.621} & \ul{30.846} &  \tb{9.268} \\
  tuhh\_day\_03 &      1.134 &      0.970 & \tb{0.731} & \ul{0.743} &      28.103 &  \tb{2.180} &  \ul{8.681} & \ul{45.054} & \tb{20.430} \\
  tuhh\_day\_04 &      0.142 & \tb{0.077} &      0.232 & \ul{0.084} &       6.505 &  \tb{2.583} &  \ul{2.698} &  \ul{6.599} &  \tb{0.903} \\
tuhh\_night\_07 &     40.566 & \ul{0.279} &      0.436 & \tb{0.227} &           x &           x &           x & \ul{46.180} &  \tb{4.085} \\
tuhh\_night\_08 &          x &      0.749 & \tb{0.685} & \ul{0.740} &           x &           x &           x & \ul{54.389} & \tb{15.067} \\
tuhh\_night\_09 &      0.103 & \tb{0.057} &      0.375 & \ul{0.094} &           x &  \tb{4.848} &           x &  \tb{5.028} &  \ul{8.429} \\
\bottomrule
\end{tabular}

\begin{tablenotes}
\small
\item *All values are in meters.
The best results are in \tb{bold}, and the second best are \ul{underlined}.
'x' denotes a divergent result. KMR and OV are shorthands for KIMERA and OpenVINS.
\end{tablenotes}

\end{threeparttable}
\end{table*}

\subsection{Lidar-Inertial SLAM} \label{sec: lidar-inertial slam}

Four SOTA open source lidar-inertial odometry methods of different paradigms are tested against MCD: LIOSAM \cite{shan2020liosam}, FLIO (Fast-LIO) \cite{xu2022fast}, DLIO \cite{chen2023direct}, and SLICT \cite{nguyen2023slict}. LIOSAM is a loosely coupled method that fuses lidar odometry and IMU data over GTSAM factor graph, with traditional plane-edge geometric features and ICP-based loop closure. FLIO is a filter-based method that uses direct feature association, and incremental global map based on the i-kd tree. DLIO is also a direct method with a key-frame-based mapping scheme. Finally, SLICT is a piecewise continuous-time optimization-based method, which uses multi-scale association and incremental global surfel map. Examples and code samples of these experiments can be found on the dataset's website.

Table \ref{tab: ATE} reports the Absolute Trajectory Error (ATE) of each method. We first notice that at least half of the sequences have an ATE above 0.5m for any method, and the ATE can vary largely even for the sequences of the same campus. 
Second, with LIOSAM, we find it performs quite well for the HSS data (the kth\_ and tuhh\_ sequences). This can be attributed to the use of an external orientation estimate from the expensive VectorNav IMU, which can be quite accurate at walking speed.
However, this advantage is lost in the high-speed sequences (ATV). Compared to LIOSAM, FLIO gives a more consistent accuracy, thanks to the robust IMU propagation and global map association, at the cost of lower accuracy due to the association strategy not leveraging edge features. DLIO has similar accuracy as FLIO, with a slightly larger computational footprint. Finally, SLICT appears to perform well and consistently in both HHS and ATV bags. This can be attributed to the continuous-time formulation, which can take care of higher speed, and the point-to-surfel association strategy on a multi-scale global map. However, SLICT can be quite computationally expensive and cannot run at high-speed lidar-rate like LIOSAM and FLIO.

\subsection{Visual-Inertial SLAM}

Three SOTA visual-inertial SLAM methods are tested against MCD: VINS \cite{qin2017vins}, KIMERA \cite{rosinol2020kimera}, and OPENVINS \cite{geneva2020openvins}. VINS is a full-fledged visual SLAM method with sliding window optimization and loop-closure + bundle adjustment. KIMERA has a VIO backbone with extra modules such as incremental mesh generation and semantic representation. OPENVINS is a filter-based method with a KLT-based feature detector and tracker, online calibration, and multi-camera fusion. All three methods are configured to run the same stereo-camera-IMU suite.

The ATE of VI SLAM methods are reported in Tab. \ref{tab: ATE}. At first glance, it can be seen that all methods struggle with the night time sequences. For VINS, all non-divergent experiments have an ATE beyond 4m, whereas KIMERA achieves better accuracy in several sequences. However, due to its additional geometric verification steps, KIMERA is more prone to losing track than VINS. OpenVINS seems to strike a better balance between robustness and accuracy, as it can complete all daytime sequences and is only compromised in nighttime HHS sequences. This compromise is primarily due to the discrete-time histogram equalization step that causes the feature to lose association in extreme lighting conditions. This result emphasizes the domain shifts and challenges in night video data, highlighting the difficulties for algorithms to adapt across diverse locations.

\subsection{Range-Aided Localization}

\begin{table*}
    \centering
    \setlength{\tabcolsep}{4pt}
    \caption{Segmentation performance of existing state-of-the-art methods on the MCD NTU (upper) and MCD TUHH (bottom) sequences. The intersection over union (IoU) is utilized as our evaluation matrix. The figure in round brackets is the relative mIoU gap on MCD ntu compared to SemanticKITTI \cite{behley2019semantickitti} (upper) and the one on MCD tuhh compared to MCD ntu (bottom), respectively. Note that ``Traffic-C'' and ``Structure-O'' are short for  ``Traffic Cone'' and ``Structure-Other'' classes, respectively.}
    \resizebox{\textwidth}{!}{
    \begin{tabular}{ m{7em} |
     m{1.5em}<{\centering} m{1.5em}<{\centering} m{1.5em}<{\centering} m{1.5em}<{\centering} m{1.5em}<{\centering} m{1.5em}<{\centering} m{1.5em}<{\centering} m{1.5em}<{\centering} m{1.5em}<{\centering} m{1.5em}<{\centering} m{1.5em}<{\centering} m{1.5em}<{\centering} m{1.5em}<{\centering} m{1.5em}<{\centering} m{1.5em}<{\centering} m{1.5em}<{\centering} m{1.5em}<{\centering} m{1.5em}<{\centering} m{1.5em}<{\centering} m{1.5em}<{\centering} m{1.5em}<{\centering} m{1.5em}<{\centering} m{1.5em}<{\centering} m{1.5em}<{\centering} | m{6em}<{\centering}}
    \hline
    Methods  & 
    \rotatebox{75}{Barrier}         & \rotatebox{75}{Bike}      & \rotatebox{75}{Building}      & \rotatebox{75}{Chair}         &
    \rotatebox{75}{Cliff}           & \rotatebox{75}{Container} & \rotatebox{75}{Curb}          & \rotatebox{75}{Fence}         &
    \rotatebox{75}{Hydrant}         & \rotatebox{75}{Sign}      & \rotatebox{75}{Lanemark}      & \rotatebox{75}{Other}         &
    \rotatebox{75}{Parkinglot}      & \rotatebox{75}{Pedestrian}& \rotatebox{75}{Pole}          & \rotatebox{75}{Road}          &
    \rotatebox{75}{Shelter}         & \rotatebox{75}{Sidewalk}  & \rotatebox{75}{Stairs}        & \rotatebox{75}{Structure-O}   &
    \rotatebox{75}{Traffic-C}      & \rotatebox{75}{Trunk}     & \rotatebox{75}{Vegetation}     & \rotatebox{75}{Vehicle}       & mIoU \\
    \hline
    MKNet \cite{choy20194d} &  
    57.5           & 59.7         & 79.9         & 34.5         &
    1.8            & 11.5         & 45.7         & 31.6         & 
    17.3           & 62.9         & 46.9         & 1.6          &
    9.2            & 48.9         & 53.3         & 89.7         &
    52.4           & 56.6         & 42.2         & 6.7          &
    8.6            & 45.1         & 87.7         & 67.5         &
    46.2 (-26.9\%)\Tstrut \\
    SalsaNext \cite{cortinhal2020salsanext} &  
    49.6           & 0.2          & 81.3         & 37.2         &
    6.4            & 0.0          & 48.7         & 36.7         & 
    9.3            & 35.5         & 45.7         & 30.2         &
    12.2           & 11.8         & 27.5         & 91.4         &
    43.7           & 66.3         & 0.0          & 0.0          &
    8.1            & 31.1         & 84.0         & 48.2         &
    33.5 (-43.6\%) \\
    SPVCNN \cite{tang2020searching} &  
    65.5           & 56.1         & 81.4         & 44.9         &
    8.6            & 16.6         & 50.6         & 37.0         & 
    22.0           & 65.8         & 49.7         & 1.3          &
    28.2           & 46.3         & 58.1         & 91.7         &
    62.9           & 93.8         & 49.1         & 6.1          &
    8.9            & 51.3         & 89.5         & 71.4         &
    48.2 (-27.5\%) \\
    Cylinder3D \cite{zhu2021cylindrical} & 
    70.5           & 1.4          & 90.1         & 42.3         &
    9.2            & 19.5         & 43.4         & 31.6         & 
    16.2           & 65.4         & 48.5         & 1.4          &
    13.6           & 41.1         & 65.1         & 90.5         &
    60.7           & 58.7         & 41.3         & 0.6          &
    10.3           & 57.2         & 91.6         & 72.1         &
    43.5 (-36.9\%) \\
    SDSeg3D \cite{li2022self} & 
    72.2           & \textbf{57.9}& 82.2         & \textbf{57.3}&
    0.8            & 0.0          & 47.6         & 31.4         & 
    24.4           & 67.1         & 49.1         & 0.0          &
    12.5           & 45.3         & 61.0         & 91.6         &
    69.0           & 64.8         & 49.0         & 0.0          &
    11.1           & 56.0         & 89.4         & 80.0         &
    46.7 (-31.4\%) \\
    WaffleIron \cite{puy2023using} & 
    76.7           & 28.3         & 81.6         & 43.9         &
    6.9            & 6.9          & 57.2         & 37.7         & 
    14.7           & 57.9         & 53.5         & \textbf{11.7}&
    39.7           & 52.5         & 56.7         & 91.5         &
    67.3           & 77.0         & 65.8         & \textbf{42.2}&
    12.4           & 59.0         & 88.2         & 75.9         &
    50.2 (-29.1\%)\\
    S-Former \cite{lai2023spherical} & 
    \textbf{79.1}  & 46.1         & \textbf{88.7}& 50.4         &
    \textbf{16.6}  & \textbf{23.1}& \textbf{61.2}& \textbf{46.7}& 
    \textbf{51.2}  & \textbf{75.4}& \textbf{55.1}& 10.4          &
    \textbf{45.1}  & \textbf{66.6}& \textbf{73.9}& \textbf{93.6}&
    \textbf{78.1}  & \textbf{77.8}& \textbf{72.6}& 37.6         &
    \textbf{26.4}  & \textbf{63.3}& \textbf{92.3}& \textbf{83.7} &
    \textbf{59.0} (-21.2\%) \\
    \hline
    \end{tabular}
    }
    \newline
    \newline
    \resizebox{\textwidth}{!}{
    \begin{tabular}{ m{7em} |
     m{1.5em}<{\centering} m{1.5em}<{\centering} m{1.5em}<{\centering} m{1.5em}<{\centering} m{1.5em}<{\centering} m{1.5em}<{\centering} m{1.5em}<{\centering} m{1.5em}<{\centering} m{1.5em}<{\centering} m{1.5em}<{\centering} m{1.5em}<{\centering} m{1.5em}<{\centering} m{1.5em}<{\centering} m{1.5em}<{\centering} m{1.5em}<{\centering} m{1.5em}<{\centering} m{1.5em}<{\centering} m{1.5em}<{\centering} m{1.5em}<{\centering} m{1.5em}<{\centering} m{1.5em}<{\centering} m{1.5em}<{\centering} m{1.5em}<{\centering} m{1.5em}<{\centering} | m{6em}<{\centering}}
    \hline
    Methods  & 
    \rotatebox{75}{Barrier}         & \rotatebox{75}{Bike}      & \rotatebox{75}{Building}      & \rotatebox{75}{Chair}         &
    \rotatebox{75}{Cliff}           & \rotatebox{75}{Container} & \rotatebox{75}{Curb}          & \rotatebox{75}{Fence}         &
    \rotatebox{75}{Hydrant}         & \rotatebox{75}{Sign}      & \rotatebox{75}{Lanemark}      & \rotatebox{75}{Other}         &
    \rotatebox{75}{Parkinglot}      & \rotatebox{75}{Pedestrian}& \rotatebox{75}{Pole}          & \rotatebox{75}{Road}          &
    \rotatebox{75}{Shelter}         & \rotatebox{75}{Sidewalk}  & \rotatebox{75}{Stairs}        & \rotatebox{75}{Structure-O}   &
    \rotatebox{75}{Traffic-C}      & \rotatebox{75}{Trunk}     & \rotatebox{75}{Vegetation}     & \rotatebox{75}{Vehicle}       & mIoU \\
    \hline
    MKNet \cite{choy20194d} &  
    0.1            & 1.6          & 69.1         & 0.0          &
    0.0            & \textbf{0.1} & 1.3          & 4.6          & 
    0.0            & 5.1          & 2.0          & 6.5          &
    4.0            & 36.7         & 31.5         & 3.3          &
    0.0            & 15.4         & 0.3          & 0.7          &
    0.0            & 29.6         & 61.9         & 37.8         &
    13.0 (-71.9\%)\Tstrut \\
    SalsaNext \cite{cortinhal2020salsanext} &  
    0.0            & 0.0          & 35.9         & 0.1          &
    0.0            & 0.0          & 0.6          & 6.2          & 
    0.0            & 3.2          & 3.1          & 0.2          &
    0.4            & 0.8          & 9.8          & 4.1          & 
    0.0            & 8.4          & 0.0          & 0.0          & 
    0.0            & 9.2          & 39.9         & 18.1         &
    5.8 (-82.1\%)\\
    SPVCNN \cite{tang2020searching} &  
    0.1            & 0.4          & 68.7         & 0.1          &
    0.1            & 0.0          & 1.7          & 5.6          & 
    0.0            & \textbf{6.4} & 2.4          & 6.8          &
    5.3            & 34.0         & 27.2         & 3.9          &
    0.0            & 17.1         & 0.5          & 0.9          &
    0.0            & 28.9         & \textbf{64.0}& 35.1         &
    12.9 (-73.3\%)\\
    Cylinder3D \cite{zhu2021cylindrical} & 
    0.1            & 0.6          & 66.8         & 0.1          &
    0.0            & 0.0          & 0.4          & \textbf{8.9} & 
    0.0            & 5.3          & \textbf{6.3} & 3.7          &
    2.1            & 14.1         & 12.0         & \textbf{6.8} &
    0.0            & 27.1         & 0.4          & 0.0          &
    0.0            & 9.2          & 50.5         & 14.6         &
    9.6 (-76.9\%) \\
    SDSeg3D \cite{li2022self} & 
    0.1            & 3.3          & \textbf{73.2}& 0.1          &
    0.0            & 0.0          & 1.1          & 8.1          & 
    0.0            & 5.3          & 2.6          & 0.1          &
    3.7            & \textbf{45.9}& \textbf{38.0}& 4.2          &
    \textbf{0.3}   & \textbf{19.9}& 1.1          & 0.0          &
    0.0            & \textbf{43.4}& 61.2         & \textbf{48.3}&
    \textbf{15.0} (-67.9\%) \\
    WaffleIron \cite{puy2023using} & 
    0.1            & 5.5          & 65.0         & 0.2          &
    0.2            & 0.0          & 1.3          & 7.9          & 
    0.0            & 6.0          & 2.5          & 0.0          &
    \textbf{5.5}   & 35.2         & 18.6         & 3.6          &
    0.0            & 8.3          & 0.1          & 1.0          &
    0.0            & 27.6         & 52.8         & 32.7         &
    11.4 (-77.3\%) \\
    S-Former \cite{lai2023spherical} & 
    \textbf{0.2}   & 1.4          & 72.8         & \textbf{0.5} &
    \textbf{0.1}   & 0.0          & \textbf{1.8} & 6.3          & 
    0.0            & 5.4          & 3.7          & 2.8          &
    3.3            & 40.1         & 33.9         & 4.5          &
    0.0            & 17.8         & \textbf{5.0} & \textbf{0.9} &
    0.0            & 35.8         & 54.8         & 40.4         &
    13.8 (-76.6\%) \\
    \hline
    \end{tabular}
    }
    \label{tab: Seg}
\end{table*}

Several SOTA range-aided localization methods with different sensor configurations are tested against MCD: RO (range-only odometry) \cite{wang2017ultra}, and VRO (visual-ranging odometry) \cite{nguyen2020tightlyauro}. Their results are shown in the last two columns in \Tab{tab: ATE}.

First, we notice that range-based methods are less likely to diverge than VIO methods because the range observations restrict the estimates to diverge too far from the anchors. However, the ATEs in many sequences are quite large compared to VIO, reflecting that existing range-based methods have only been tested in small environments with consistent LOS (line-of-sight) to all anchors. In MCD, UWB anchors are deployed sparsely over the kilometer-long routes. Thus, the robot can lose LOS to some anchors frequently. In contrast, thanks to the visual factors filling in when UWB LOS is lost, the ATE is much better with the VRO method. Nevertheless, the accuracy can still be improved with more robust algorithms.

\subsection{Semantic Segmentation} \label{sec: semantic segmentation}

3D semantic segmentation is a long-standing topic that has attracted increasing attention in recent years. While existing methods are mostly tested on point clouds captured by cylindrically sweeping lidar, we investigate whether their state-of-the-art performance can be maintained on point clouds from NRE lidar. 

\noindent \textbf{Baselines.} Eight SOTA methods for 3D semantic segmentation are tested on MCD, including SalsaNext \cite{cortinhal2020salsanext}, MKNet (MinkowskiNet) \cite{choy20194d}, SPVCNN \cite{tang2020searching}, SDSeg3D \cite{li2022self}, Cylinder3D \cite{zhu2021cylindrical}, WaffleIron \cite{puy2023using}, and S-Former (SphereFormer) \cite{lai2023spherical}. Specifically, SalsaNext is a projection-based method utilizing a CNN-based backbone with faster inference speed, which introduces an additional residual dilated convolution module compared to the previous range-image-based method \cite{milioto2019rangenet++}. On the other hand, MKNet, SPVCNN, and SDSeg3D are regarded as voxel-based methods that leverage sparse convolution \cite{choy20194d} as their main feature extractor given 3D voxels as input. SPVCNN \cite{tang2020searching} further presents an efficient point-voxel feature fusion to embed dense point-wise features while SDSeg3D introduces self-distillation and test-time augmentation for performance boost. The rest of the baseline methods consider various 3D representations to cope with the 3D pattern of the raw input, such as cylindrical partitions (Cylinder3D \cite{zhu2021cylindrical}), spherical partitions (SphereFormer \cite{lai2023spherical}), and high-level point-wise representations (WaffleIron \cite{puy2023using}).

\noindent \textbf{Results and Analysis.} All baseline methods are trained on the NTU subset of MCD with fully supervised learning. As presented in the upper table of Tab.~\ref{tab: Seg}, while training and testing within the same campus, existing 3D segmentation methods can not adapt well to the different point cloud patterns of NRE lidar, where all baseline methods suffer from serious mIoU drops of more than $25\%$ relatively compared to training on SemanticKITTI. For methods based on cylindrical-specific 3D representation, such as Cylinder3D with cylindrical partitions and projection-based SalsaNext, the gaps are more severe since their designed representations are ineffective due to the different point cloud patterns. Voxel-based methods, including MKNet, SPVCNN, and SDSeg3D, also suffer from a significant performance drop mainly caused by point density discrepancy. Among all baseline methods, WaffleIron and S-Former are the only two approaches that achieve more than $50\%$ on mIoU, mainly because their pre-defined representation can be propagated to point clouds captured by NRE lidar. Nevertheless, they similarly suffer from poor segmentation performance on some rare but important objects, such as traffic cones (``Traffic-C'') and hydrants. 

When testing on a different campus MCD TUHH as in the bottom table of Tab.~\ref{tab: Seg}, the performance of all methods significantly decreases by relatively more than $70\%$ compared to the scores they achieve on MCD NTU, which indicates that the existing methods seriously suffer from the cross-campus discrepancy. In fact, this cross-campus inference task can be viewed as a \textit{domain adaptation} problem \cite{peng2021sparse,cao2023multi}, where the domain gap is mainly attributed to the infrastructure difference between different campuses and different weather conditions. It is worth further investigation to improve the generalizability of existing segmentation networks. Detailed experimental settings and further analysis are provided in our supplementary materials.

\subsection{Discussion and Insights}

Based on our experiment benchmarks, no single SOTA SLAM method consistently excels across all MCD sequences. Additionally, there's a noticeable gap in SOTA semantic segmentation research for the NRE lidar modality. This shows the complexity of MCD through several diverse scenarios and conditions. It is our hope that the challenges in MCD can inspire new accurate and robust methods covering a variety of environments. Our configurations for these experiments will be shared on the MCD website.

\section{Conclusion}

This work introduces MCD data suite, featuring a wide range of the latest sensors and extensive coverage of diverse campuses across the Eurasian continent, along with semantic annotation on NRE lidar scans, and novel high-accuracy continuous-time ground truth. Tested with various state-of-the-art SLAM and perception methods, the dataset uncovers numerous challenges, calling for robust and precise solutions from the research community.

\textbf{Acknowledgements} This research is supported by the National Research Foundation, Singapore under its Medium Sized Center for Advanced Robotics Technology
Innovation, and the Wallenberg AI, Autonomous Systems and Software Program via the 2020 Wallenberg-NTU Presidential Postdoctoral Fellowship.

\end{document}

%% file: preamble.tex
\usepackage[dvipsnames]{xcolor}

\usepackage{times}
\usepackage{amssymb}

\usepackage[utf8]{inputenc}

\usepackage{mathbbol}

\usepackage{esvect}

\usepackage[percent]{overpic}

\usepackage{tabu}

\usepackage{multirow}
\def\mr{\multirow}

\usepackage[overload]{empheq}

\usepackage{bm}

\definecolor{dg}{rgb}{0.1, 0.6, 0.2}       
\definecolor{b}{rgb}{0.0, 0.0, 1}          

\usepackage{epsfig}
\usepackage{float}
\usepackage{color}
\usepackage{booktabs} 
\usepackage{multirow} 
\usepackage{algpseudocode}
\usepackage[ruled,linesnumbered,vlined]{algorithm2e}

\usepackage{amsfonts}
\usepackage{amsmath}
\usepackage{mathrsfs}

\usepackage{amsthm}
\usepackage{mathtools}

\usepackage{enumitem}       
\setenumerate[enumerate]{align=left}

\usepackage{graphicx}
\usepackage{subcaption}
\usepackage{caption}

\usepackage{balance}

\usepackage{multirow}
\def\mr{\multirow}

\usepackage[flushleft]{threeparttable}

\newcommand{\norm}[1]{\left\lVert#1\right\rVert}

\makeatletter
\newlength\tmp@\newlength\t@mp
\newcommand{\comp}[3]
  {\mathop{ \settowidth\tmp@{$\displaystyle\mathop{#1}^{#3}_{#2}$}
  \hbox to \tmp@{\hss \settowidth\t@mp{$\displaystyle #1$}\setlength\t@mp{.45\t@mp}
  $\displaystyle\mathop{#1}^{\hspace\t@mp #3}_{\hspace{-\t@mp}#2}$
  \hss} }}
\makeatother

\newcommand{\Int}[2]
{\int_{#1}^{#2}}

\def\ll{\left}
\def\rr{\right}

\def\pos{\mathbf{p}}

\newcommand{\fr}[1]{\texttt{#1}}

\newcommand{\vbf}[1]{{\bm{\mathbf{#1}}}}

\def\bias{\mathbf{b}}

\def\rot{\mathbf{R}}
\def\tf{\mathbf{T}}

\def\Log{\mathrm{Log}}

\def\cov{\mathrm{G}}

\def\Z{\mathcal{Z}}

\def\fB{\fr{B}}

\def\f{\vbf{f}}
\def\n{\vbf{n}}

\def\ssp{\text{pose}}
\def\ssl{\text{lidar}}
\def\ssg{\text{gyro}}
\def\ssa{\text{acce}}

\def\fB{\fr{B}} 

\newcommand{\ul}[1]{\underline{#1}}

\newcommand{\tb}[1]{\textbf{#1}}

\newcommand{\Fig}[1]{Fig. \ref{#1}}

\newcommand{\Tab}[1]{Tab. \ref{#1}}

\usepackage{xstring}
\newcommand{\shorturl}[1]
{
    \href{https://#1}{\nolinkurl{#1}}
}

\usepackage{array}
\newcommand\Tstrut{\rule{0pt}{2.2ex}}         